\pdfoutput=1
\documentclass[11pt]{article}

\usepackage[final]{coling}

\usepackage{times}
\usepackage{latexsym}

\usepackage[T1]{fontenc}
\usepackage[utf8]{inputenc}

\usepackage{microtype}

\usepackage{inconsolata}

\usepackage{graphicx}
\usepackage{hyperref}
\usepackage{amsmath}
\usepackage{anyfontsize}
\usepackage{booktabs}
\usepackage{float}
\title{From Superficial to Deep: Integrating External Knowledge for Follow-up Question Generation Using Knowledge Graph and LLM}

\author{
 \textbf{Jianyu Liu\textsuperscript{1}},
 \textbf{Yi Huang\textsuperscript{3}\thanks{Corresponding author}},
 \textbf{Sheng Bi\textsuperscript{2}},
 \textbf{Junlan Feng\textsuperscript{3}},
 \textbf{Guilin Qi\textsuperscript{1}},
\\
\\
 \textsuperscript{1}School of Computer Science and Engineering, Southeast University, China
\\
 \textsuperscript{2}Law and Innovation Lab, Law School, Southeast University, China
\\
 \textsuperscript{3}China Mobile Research Institute, China
\\
 \small{
   \textbf{Correspondence:} {\{liujianyu, bisheng, gqi\}@seu.edu.cn}, {\{huangyi, fengjunlan\}@chinamobile.com}
 }
}

\begin{document}
\maketitle
\begin{abstract}
In a conversational system, dynamically generating follow-up questions based on context can help users explore information and provide a better user experience. Humans are usually able to ask questions that involve some general life knowledge and demonstrate higher order cognitive skills. However, the questions generated by existing methods are often limited to shallow contextual questions that are uninspiring and have a large gap to the human level. In this paper, we propose a three-stage external knowledge-enhanced follow-up question generation method, which generates questions by identifying contextual topics, constructing a knowledge graph (KG) online, and finally combining these with a large language model to generate the final question. The model generates information-rich and exploratory follow-up questions by introducing external common sense knowledge and performing a knowledge fusion operation. Experiments show that compared to baseline models, our method generates questions that are more informative and closer to human questioning levels while maintaining contextual relevance.
\end{abstract}

\section{Introduction}
Asking questions is a fundamental way for humans to learn new knowledge. Question generation (QG), an important task in the field of natural language processing, aims to generate a question based on a given text. A good question is crucial for a conversational system, because an excellent system should be able to interact well with the user through asking and responding~\cite{ref1}. With the rapid development of artificial intelligence technology, generative AI conversational systems have been widely used in many fields, such as education~\cite{ref22,ref25}, healthcare~\cite{ref24}, and legal consultation~\cite{ref26}. However, while large language models (LLM) such as ChatGPT~\cite{gpt3.5} can respond to queries, they are often passive - only responding to user queries rather than proactively guiding the conversation or posing their own inquiries. To address this limitation in proactivity, the task of follow-up QG was introduced~\cite{ref17}.

\begin{figure}[t]
  \centering
  \includegraphics[width=0.45\textwidth]{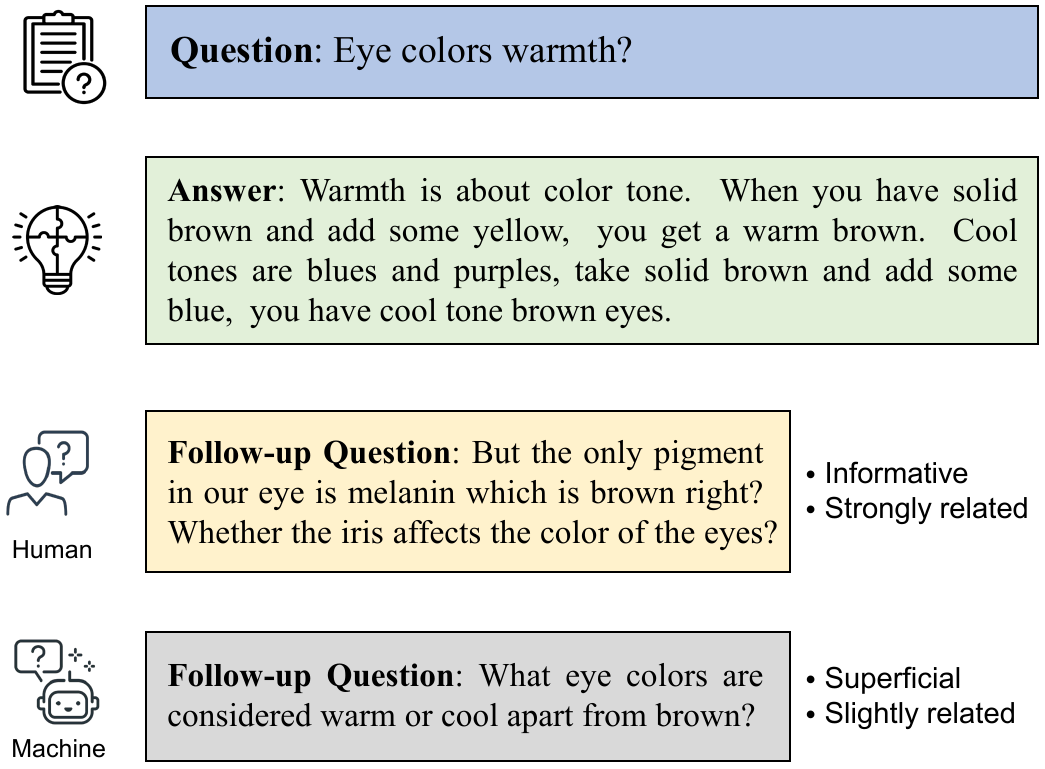} 
  \caption{When humans ask questions, they can rely on relevant common knowledge to introduce new directions of thought. However, it is difficult to achieve with existing methods.}
  \label{fig1}
\end{figure}

In a conversational system, a follow-up question usually refers to a continuation question generated based on the user's input or the system's initial answer~\cite{ref16}. Such questions differ significantly from those produced by traditional question generation tasks~\cite{ref29}, where the generated question can be answered using the source text. In contrast, a follow-up question cannot be answered in the previous context. Intuitively, a good follow-up question must meet two requirements while maintaining coherent and fluent formulation: (1) Ensure contextual relevance. The question should be highly relevant to the current dialog topic and should not deviate from the previous conversation content; (2) Aim to explore new information, thereby guiding the next response to provide more novel information and advance the dialog to a deeper level. 

Meng et al.~\shortcite{ref2} found that machines struggle to generate relevant questions by integrating background knowledge and examples, resulting in a significant gap in the amount of information compared to humans. Moreover, humans can generate follow-up questions through higher-level cognitive skills, such as using analogy and association~\cite{ref27}. Pan et al.~\shortcite{ref29} suggested that due to the limitations of training data and preset models, machine-generated questions mostly remain at the level of surface linguistic relevance, lacking flexibility and creativity. An intuitive example is shown in Figure~\ref{fig1}. When discussing eye color, humans can associate other factors that are not mentioned in the context, such as melanin and the iris. However, since the machine can only rely on contextual information, the generated questions, while relevant to the previous context, often lack sufficient depth and breadth in their content.
%However, since machines can only use the context, the generated question are related to the historical conversation, they lack depth and breadth.

% The challenges of follow-up question generation are mainly in two aspects:(1)Lack of contextual background knowledge: Humans are able to ask relevant questions by providing new, concrete and relevant examples, but models lack relevant background knowledge. (2)Low cognitive level: Human beings generate follow-up questions through high-level cognitive ability, such as analogy, correlation, etc.. However, empirical studies~\cite{ref29} show that due to the limitations of training data and preset models, machine-generated questions mostly remain at the level of surface linguistic relevance, lacking flexibility and creativity. An intuitive example is shown in Figure~\ref{fig1}. When discussing eye color, humans can associate other factors that are not mentioned in the context, such as melanin and iris. However, since machines can only use the context, although the generated question are related to the historical conversation, they lack depth and breadth.

In this paper, we address the above limitations and propose a method that introduces external knowledge through the online construction of KG and combines it with LLM to generate the follow-up question. Specifically, we first perform intent recognition on historical question and answer information to expand relevant background knowledge, extract core keywords from the conversation, and construct a query to retrieve the most relevant Wikipedia page. Next, we construct a real-time KG centered on the entity corresponding to the page. We then select the nodes most relevant to the conversation based on two dimensions: node importance and relevance, thereby identifying the external background knowledge to be introduced. This allows the model to access a broader range of knowledge resources, improving the depth and relevance of the generated question. To address the challenge of the model's limited cognitive ability, we design a knowledge fusion operation to further enhance the model's understanding and cognition of the context by instructing the LLM to continue writing the previously acquired external Wikipedia knowledge based on the context. In summary, our contributions are as follows:

\begin{itemize}
    \item We develop a three-stage framework for follow-up question generation, integrating multi-source knowledge to generate coherent, clear and informative follow-up questions.
    \item We design a strategy to inject common sense knowledge into the question generation process by constructing a KG online, making the questions more knowledge-supported.
    \item We conduct extensive experiments and analysis, demonstrating the superiority of the proposed method in the this task.
\end{itemize}

\section{Related Work}
Question generation aims to automatically generate semantically reasonable and structurally complete questions from a given text~\cite{ref28}. Traditional question generation has been widely applied in fields such as machine comprehension~\cite{ref18,ref19}, e-commerce~\cite{ref20,ref21}, educational guidance~\cite{ref22}, news media~\cite{ref23} and other fields. In these tasks, the answers to the generated questions are known as they derive from the information provided to the model. This is fundamentally different from the starting point of human questioning, which is driven by the search for new information. In this work, we aim to generate a follow-up question that probes for unknown information within the given knowns.

Previous work focusing on follow-up question generation mainly concentrated on rule-based or using pre-trained language models (PLM). Template filling ~\cite{ref12,ref13} not only limits the diversity of question types but also fails to generate personalized questions. Kumar and Joshi~\shortcite{ref14} proposed a sequence-to-sequence retrieval-based learning system to generate complete questions for incomplete follow-up questions. Su et al.~\shortcite{ref15} focused on applications in interview systems, using a pattern-based sequence-to-sequence model for follow-up question generation. Ge et al.~\shortcite{ref16} proposed a knowledge-driven framework for follow-up question generation, combining a knowledge selection module and a generation model guided by selected knowledge entity-relation pairs. Wang et al.~\shortcite{ref17} designed two types of decoders (soft type decoders and hard type decoders) to generate questions by estimating the type distribution of the question components. However, with the development of LLM, there is a lack of methods for generating follow-up questions based on LLM. Meng et al.~\shortcite{ref2} found that both PLM and LLM-generated questions are far from human-asked questions in terms of information content and complexity, indicating that this task is still quite challenging.

\section{Methodology}
\begin{figure*}[t]
    \centering
    \includegraphics[width=0.98\linewidth]{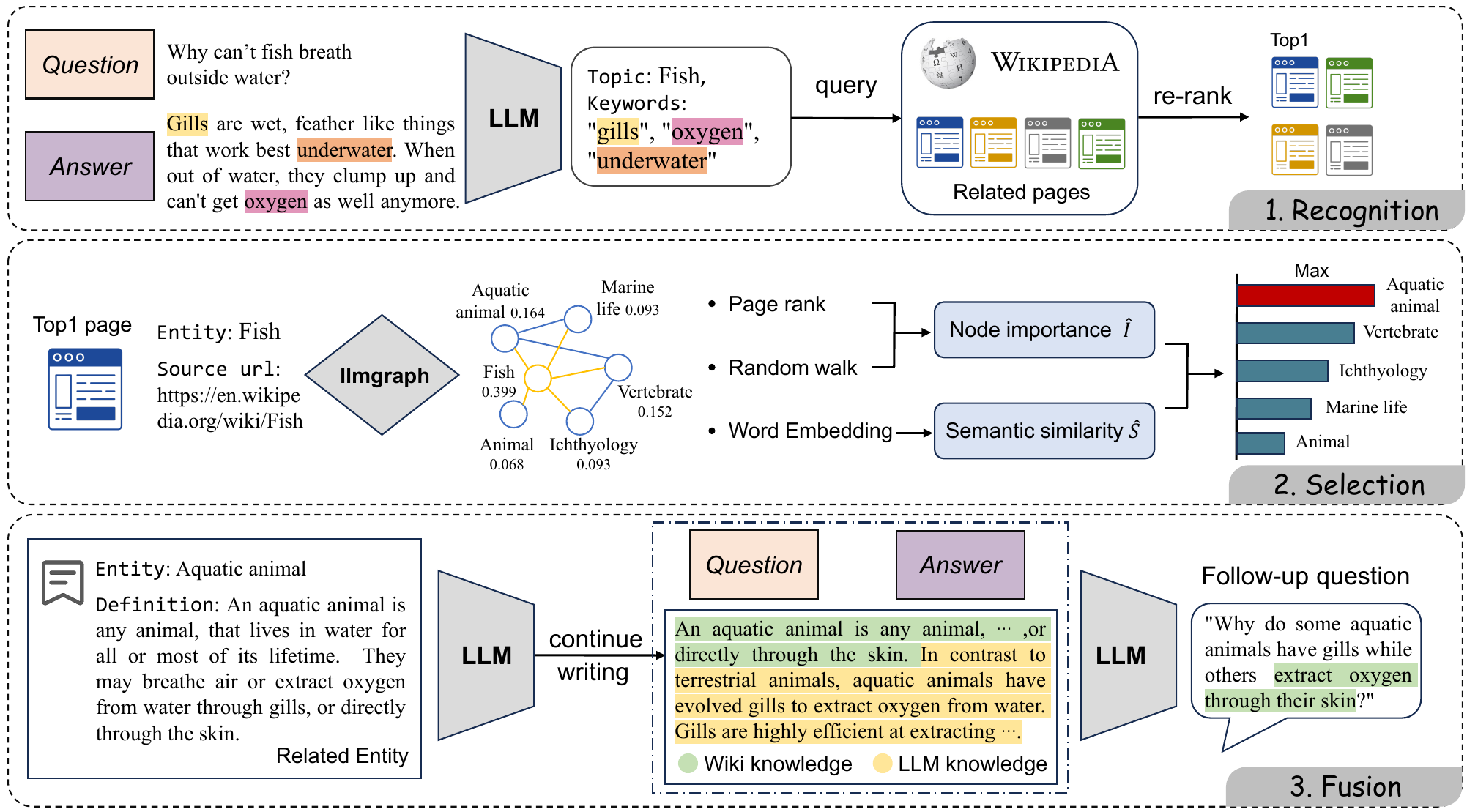}
    \caption{For a Q\&A pair in a user conversation, we first identify the key information of the context. Then, we construct a KG and select the node that is most relevant to the dialog. Finally, we integrate the background knowledge into the question generation process to generate the final follow-up question.}
    \label{fig2}
\end{figure*}

Follow-up question generation (QG) is to generate questions based on the dialogue context, aiming to steer the conversation toward a deeper level and higher creativity, as shown in Figure~\ref{fig1}. The method proposed in this paper aims to introduce external background knowledge related to the context by constructing a KG to compensate for the shortcomings of previous work that relies only on the surface information of the context to generate question. In addition, we use the knowledge fusion operation to enhance the cognitive level of the generated question, which helps to lead the conversation to a deeper level. Our framework is divided into three stages, namely \textit{Recognition}, \textit{Selection}, and \textit{Fusion}, as shown in Figure \ref{fig2}. 
% The reasons for the framework design can be summarized as follows:

\subsection{Recognition}
The primary characteristic of a follow-up question is contextual relevance. Thus, the goal of the \textit{Recognition} module is to identify the core topic of the historical dialog and extract the correct contextual information for generating the follow-up question.

First, for a given question-answer (QA) pair, we input it into the LLM and instruct it to extract the key information from the QA pair. During the extraction process, the LLM extracts one word as the \texttt{Topic} and $n$ \texttt{Keywords}. The purpose of this is that the \texttt{Topic} is a highly concise description of the entire QA, capturing the user's main question intention, while the \texttt{Keywords} extract more fine-grained details from the QA. We consider that the combination of the \texttt{Topic} and \texttt{Keywords} can better model the overall dialog.

Wikipedia\footnote{\url{https://www.wikipedia.org/}} is a multilingual online encyclopedia whose content covers almost all known fields. We query Wikipedia based on the \texttt{Topic} and \texttt{Keywords} to obtain related contextual entities in a iterative retrieval way. Specifically, we first query pages whose titles contain the \texttt{Topic}, which is denoted as the set $\mathcal{C}$. Then, we traverse the \texttt{Keywords} one by one, adding each keyword to the query condition and retrieving pages from $\mathcal{C}$ that contain the keyword until all keyword traversals have been completed. By narrowing down the query range, we obtain the most relevant related pages. During the dynamic search process, if a unique match result is found, the search process is terminated early\footnote{In this paper, we use Elasticsearch as the specific implementation of the search engine.}.

Since word overlap scores from search engines may not accurately reflect text relevance, we introduce a re-ranker to re-rank related pages. Inspired by the work of ~\cite{ref3}, the re-ranker in the \textit{Recognition} module measures the relevance of a page by calculating the following probability:
\begin{equation}
    \begin{aligned}
        P(Q \mid p_i)=\frac{1}{|Q|} \prod_{t} P(Q_t \mid Q_{<t}, p_i; \Theta),
    \end{aligned}
\end{equation}
where $Q$ is a query consisting of a concatenation of \texttt{Topic} and \texttt{Keywords}, and $\Theta$ represents the parameters of PLM. $p_i$ refers to the definition of the entity in each page. The conditional probability of the entire output sequence $Q$ is computed by the product of the conditional probabilities of all time steps, where \( P(Q_t \mid Q_{<t}, p_i; \Theta)\) is the conditional probability that the model generates the current token at each time step, which can be expressed as:
\begin{equation}
    \begin{aligned}
        P(Q_t \mid Q_{<t}, p_i; \Theta) = \textit{Softmax}(f(Q_{<t}, x; \Theta)),
    \end{aligned}
\end{equation}
among them, \( f(Q_{<t}, p_i; \Theta) \) is the output of the model, which represents the score of generating the current token \( Q_t \), which is transformed into a probability distribution through the softmax function. In this work, we choose T5~\cite{ref4} as the base model of the re-ranker. After the re-ranking is completed, the first ranked page is considered as the topic entity capable of representing the historical dialog.

\subsection{Selection}
After obtaining the page output from the \emph{Recognition} module, we use llmgraph~\footnote{\url{https://github.com/dylanhogg/llmgraph}} to construct a KG for the candidate page. The llmgraph is an open-source tool that utilizes LLM to construct KGs, capable of creating a KG from the Wikipedia page of the given source entity. We provide the URL of the page output by the \emph{Recognition} module to llmgraph, and specify the entity of the page as the central node of the resulting KG. Since the other nodes of the output KG are all associated with the central node, and each node stores the definition of its corresponding entity, the KG thus reflects the background knowledge relevant to the context.

Furthermore, we need to evaluate the relevance of these entities in order to select meaningful background knowledge. The evaluation dimensions include node importance and semantic similarity. First, we need to prioritize the most relevant and well-known entities among the many candidates, ensuring that the background knowledge introduced is consistent with common sense. We use PageRank~\cite{ref30} to analyze the overall structure in the graph and assign weights $w_i$ to each node $V_i$. Entities with global importance are given higher weights. Then, by executing the random walk~\cite{ref35} and recording the number of visits to node $V_i$, denoted as $n_i$, we calculate the importance score of the node $V_i$ as follows:
\begin{equation}
    \begin{aligned}
        I_i=w_i \times n_i,
    \end{aligned}
\end{equation}
in addition, it is necessary to ensure the relevance of the introduced background knowledge to the context. We use BERT~\cite{ref5} to encode the query in \emph{Recognition} and the definition of each entity separately, denoted as $q$ and $e_i$ respectively. The semantic similarity between the two is then computed as follows:
\begin{equation}
    \begin{aligned}
        S_i=\frac{q\cdot e_i}{\|q\|\|e_i\|},
    \end{aligned}
\end{equation}
as $S_i$ ranges from -1 to 1, we perform a max-min normalization on $I_i$ to balance the influence of $I_i$ and $S_i$. The normalization is calculated as follows:
\begin{equation}
    \begin{aligned}
        I^{'}_{i}=\frac{I_i-I_{\min}}{I_{\max}-I_{\min}},
    \end{aligned}
\end{equation}
where $I_{\max}=\max\{I_1,I_2,... ,I_n\}$, $I_{\min}=\min\{I_1,I_2,... ,I_n\}$. By combining the node's importance score $I^{'}{_i}$ and the semantic similarity $S_i$, the final composite score $R_i$ for each entity is:
\begin{equation}
    \begin{aligned}
        R_i=I^{'}_{i}+\beta\times S_i,
    \end{aligned}
    \label{beta_value}
\end{equation}
where $\beta$ is the weighting factor to balance the effect between node importance and semantic similarity. We select the entity with the highest score, and the corresponding definition is used as the introduced Wiki knowledge to provide meaningful background knowledge for generating a follow-up question to improve accuracy and reliability.

\subsection{Fusion}
LLM acquire enormous knowledge from massive text corpora. Petroni et al.~\cite{ref6} pointed out that in addition to learning linguistic knowledge, LLM also learn a significant amount of world knowledge (or factual knowledge). Unlike traditional knowledge bases where information is stored explicitly, in these LLM, knowledge is embedded within the parameters. It is necessary to guide the model appropriately to "speak out" the knowledge. Inspired by existing works~\cite{ref7,ref8}, we adopt prompt learning to try to extract knowledge from the LLM. We innovatively designed a text continuation task, requiring the model to continue writing the Wiki knowledge output by the \emph{Selection} module based on the context. The purpose of this is twofold: on one hand, to stimulate the LLM to integrate its internal world knowledge and provide more common sense knowledge; on the other hand, to integrate the knowledge with the context through natural language generation to ensure that the generated question has strong contextual relevance.

For the continue writing prompt, we designed it as follows: \texttt{\small ``Given a question-answer pair: [\textit{Question}], [\textit{Answer}]. Please continue writing the following sentences with a few sentences based on the question-answer pair to reflect the association with it.''}. During the continue writing process, LLM conducts in-depth analysis of the context and Wiki knowledge, which improves the level of understanding of background knowledge.

Finally, we instruct the LLM to generate a follow-up question based on the known information. To obtain high-quality follow-up question, we provide clear task description and context in the prompt, including the question, answer, background knowledge fragment, and language style requirement. The specific design is as follows: \texttt{\small ``Given the following information: [\textit{Question}], [\textit{Answer}], [\textit{Related knowledge}]. Based on this information, raise a follow-up question that is relevant to the question-answer content and that is thoughtful''}. At this point, we obtain the final follow-up question.

\section{Experiment Setup}
\textbf{Dataset} To evaluate the effectiveness of the proposed method, we utilize FOLLOWUPQG~\cite{ref2} as the experimental dataset. The source of FOLLOWUPQG comes from the Reddit subforum Explain Like I'm Five (ELI5), where the questions are close to real-life scenarios, and the answers are self-contained, thus requiring minimal prior expertise. FOLLOWUPQG contains 3,790 samples, each structured as a triplet consisting of an initial question, an answer, and a follow-up question. Since all data are derived from human responses, FOLLOWUPQG captures a variety of high-level cognitive skills in the questions, such as association and causal reasoning.

\noindent \textbf{Evaluation Metrics} In the experiment, we report a range of representative metrics relevant to the task to assess the quality of the generated results, including Topic Consistency, Mutual Information (MI)~\cite{ref31}, Distinct-$n$~\cite{ref9}, and Type-Token Ratio (TTR)~\cite{ref32}, which respectively reflect relevance, informativeness, and diversity, with Distinct-$n$ and TTR both capturing aspects of diversity. For Topic Consistency, we use LDA~\cite{ref33} to extract the topics from the conversation context and the follow-up question. We select the top $N$ words with the highest probability for each topic, and calculate their co-occurrence frequency for scoring. The average score across all topics determines the consistency between context and questions, with a higher score indicating a stronger thematic correlation. MI measures how much the uncertainty of one variable is reduced given the value of another variable. In the experiment, we compute the MI between the follow-up question and the initial question. When MI is smaller, the initial question reveals less information about the follow-up question, meaning that the generated question contains more information. Existing research suggests that diversity is a reliable basis for measuring the creativity of generated content~\cite{ref36,ref37}.

\noindent \textbf{Baselines} We select the following baseline models for comparison:
\begin{itemize}
    \item \textbf{PLMs}: We use the baseline model set by the FOLLOWUPQG dataset for comparison, including BART~\cite{ref10}, T5~\cite{ref4}, and GPT-Neo~\cite{ref11}. All models are fine-tuned on the training set.
    % We fine-tune the BART~\cite{ref10} and T5~\cite{ref4} on the training set, taking the concatenation of the initial question and the answer as input, and the model predicts the output follow-up question. When fine-tuning GPT-Neo~\cite{ref11}, we use the concatenation of the initial question, answer, and follow-up question as the input sequence, and when testing, only the initial question and answer are provided.
    \item \textbf{LLMs}: We introduce several mainstream open-source and closed-source LLMs for comparison, including gpt-3.5-turbo~\cite{gpt3.5}, LLaMA3~\cite{llama3}, Qwen2~\cite{qwen2}, and ChatGLM4~\cite{chatglm}. We use the standard prompt to instruct each model to complete the task.
\end{itemize}

\noindent \textbf{Implementation Details} For the configuration of our method, the number of extracted keywords $n$ in the \emph{Recognition} module is set to 3, the number of random walk steps in the \emph{Selection} module is set to 100, and the embedding model used to measure semantic similarity is all-MiniLM-L6-v2. The $\beta$ value in equation~\ref{beta_value} is set to 1.0, and all LLM used are gpt-3.5-turbo. For LLM deployment, we use vLLM~\cite{vllm} to accelerate the inference, with the temperature set to 1.0. All experiments are conducted on a cluster of NVIDIA 4090 24GB GPUs.

\section{Experimental Results}
\subsection{Main Results}
\begin{table*}[ht]
    \centering
    \resizebox{\textwidth}{!}{
    \renewcommand{\arraystretch}{1.1}
    \begin{tabular}{cccc|ccccc}
        \toprule
          & \multicolumn{3}{c|}{\fontsize{10}{10} \selectfont \textbf{Pre-trained language model-based}} & \multicolumn{5}{c}{\fontsize{10}{10} \selectfont \textbf{Large language model-based}} \\ \cline{2-9}
         & BART & T5 & GPT-Neo & LLaMA3 & Qwen2 & ChatGLM4 & gpt-3.5-turbo & Ours \\ \hline
        Consistency(\%) & 62.11 & 61.13 & \textbf{77.11} & 54.16 & 53.83 & 54.15 & 52.74 & 54.42 \\ \cline{1-1}
        Mutual Information & 0.7850 & 0.7535 & 1.2349 & 0.7816 & 0.7943 & 0.7921 & 0.7677 & \textbf{0.7515} \\ \cline{1-1}
        Distinct-1(\%) & 30.57 & 28.11 & 18.61 & 31.63 & 31.43 & 30.21 & 31.73 & \textbf{33.84} \\ \cline{1-1}
        Distinct-2(\%) & 70.64 & 61.59 & 68.06 & 70.46 & \textbf{70.79} & 70.33 & 68.91 & 67.88 \\ \cline{1-1}
        TTR(\%) & 92.70 & 86.00 & 65.65 & 92.61 & 93.51 & 92.16 & 96.65 & \textbf{97.08} \\ \toprule
    \end{tabular}
    }
    \caption{Experimental results of various methods on the follow-up question generation task.}
    \label{table1}
\end{table*}

\begin{figure*}[t]
  \centering
  \begin{minipage}[b]{0.6\textwidth} % 左边放表格
      \centering
      \resizebox{\textwidth}{!}{
      \renewcommand{\arraystretch}{1}
      \setlength\tabcolsep{1.5pt}
          \begin{tabular}{lcccccc}
          \toprule
          \textbf{Models} & \textbf{BART} & \textbf{LLaMA3} & \textbf{Qwen2} & \textbf{ChatGLM4} & \textbf{gpt-3.5-turbo} & \textbf{Ours}  \\
          \midrule
          \textbf{COM.}    & \large 0.08 & \large 0.21 & \large 0.24 & \large 0.19 & \large 0.21 & \large \textbf{0.42}  \\
          \textbf{REL.}    & \large \underline{0.98} & \large  0.97 & \large 0.97 & \large 0.96 & \large \underline{0.98} & \large 0.97  \\
          \textbf{INF.} & \large 1.24 & \large 1.68 & \large 1.76 & \large 1.71 & \large 1.66 & \large \textbf{1.98}  \\
          \bottomrule
          \end{tabular}
      }
      \captionof{table}{Performance of each model on human evaluation for follow-up question generation. \textbf{COM.}: Complexity (0-1); \textbf{REL.}: Relevance (0-1); \textbf{INF.}: Informativeness (1-3).}
      \label{table3}
  \end{minipage}%
  \begin{minipage}[b]{0.4\textwidth} % 右边放图片
      \centering
      \includegraphics[width=0.5\textwidth]{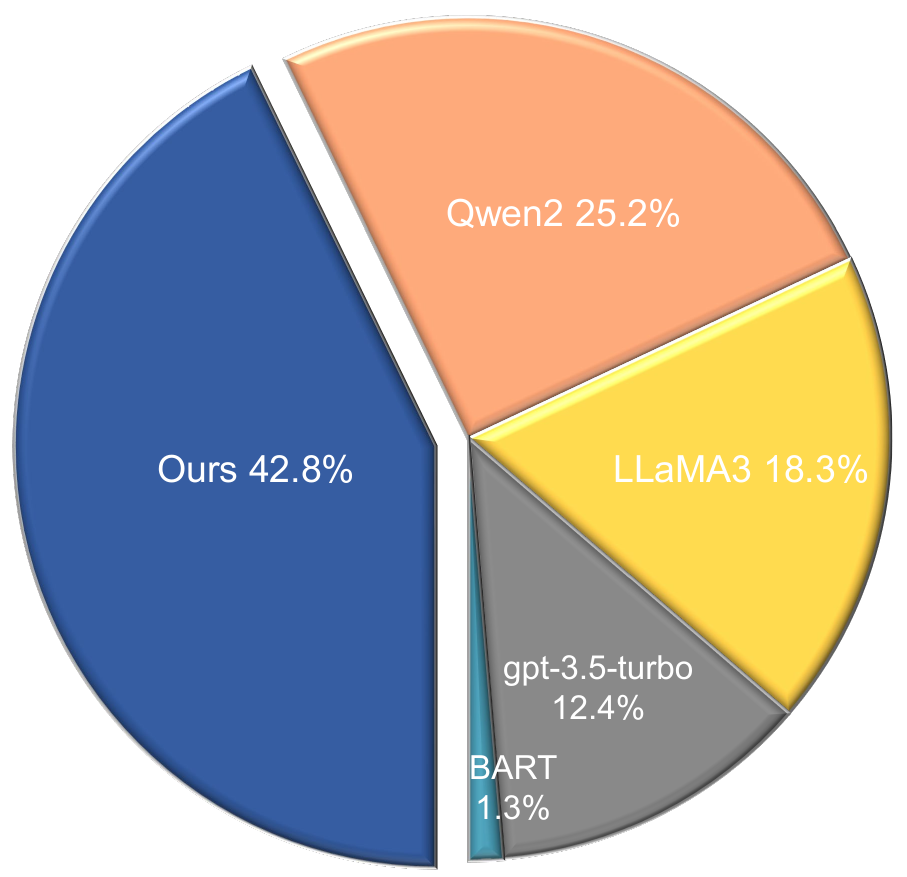}
      \caption{User preference distribution.}
      \label{fig3}
  \end{minipage}
\end{figure*}

We report the comparative results of various models on the FOLLOWUPQG dataset~\cite{ref2} in Table~\ref{table1}. Regarding the relevance between the generated responses and the context, we observe that PLMs generally outperform than LLMs in terms of Topic Consistency. A direct reason for this is that PLMs have a much smaller parameter size compared to LLMs, making smaller models more prone to overfitting on the training data. As a result, PLMs tend to paraphrase the phrases from the input context. In such cases, the question generated by PLMs is closer to the input content, resulting in higher Topic Consistency. Therefore, the higher Topic Consistency of PLMs does not necessarily indicate better question quality. Compared to other LLMs, thanks to the extraction of key contextual information by the \emph{Recognition} module, our method generates question that is more related to the context, effectively maintaining topic consistency with the input.

Our method achieves the lowest MI, i.e. the initial question reveals the least amount of information about the generated follow-up question. It is noteworthy that T5 also achieves relatively low MI. Further investigation reveals that T5 tends to generate question texts that contain many meaningless metacognitive expressions, such as "This is baffling to me," "I'm not doubting you," or "I'm not sure...". These expressions are modal expressions at the discourse level, but they are irrelevant to the actual question and can be considered as redundant filler words. The proportion of such expressions exceeds 17\% and contributes to the low MI of T5. In contrast, our method does not exhibit this phenomenon. By leveraging a \emph{Selection} module that constructs a KG and selects the most contextually relevant and widely known entities from KG as external knowledge, our method generates follow-up questions that do not simply repeat or rephrase the original context, but introduce new information to initiate new topics. Additionally, the \emph{Fusion} module integrates Wiki knowledge and LLM knowledge, ensuring contextual relevance while allowing for deeper exploration, thus achieving a good balance between the two objectives.

Our method also outperforms baseline models in terms of text diversity. We achieve the highest Distinct-1 and TTR, indicating that the injection of external knowledge contributes to more diverse follow-up questions. Among the baseline models, LLMs generally outperform than PLMs, which suggests that LLMs, trained on larger knowledge bases, generate more diverse texts. In comparison to other LLMs, our method showed improvements in all of the evaluation metrics, particularly when compared to the base model used in the framework, gpt-3.5-turbo, which further validates the effectiveness of our framework design.

\subsection{Human Evaluation}
We employ crowdsourcing to perform a human evaluation of 100 randomly selected samples from the FOLLOWUPQG test set. Five English-proficient workers are asked to evaluate the questions generated by different models for a specific sample. The detailed criteria are shown in the questionnaire provided in Appendix~\ref{appendix1}. For PLMs, we only selected BART that performed best in the automatic evaluation. Workers are blinded to the identity of the model to which the question belongs. For each question, we ask workers to score it based on three criteria: \textit{Complexity, Relevance}, and \textit{Informativeness}. Then workers are asked to choose the question they would most like to ask from all the questions. We count and average the scores on each question and report the average performance, which is shown in Table~\ref{table3} and Figure~\ref{fig3}.

According to Table~\ref{table3}, all models can maintain contextual relevance effectively, but our method is significantly better than the other models in \emph{Complexity} and \emph{Informativeness}, particularly with at least an 18\% improvement in \emph{Complexity} over the baselines. It indicates that our method can use a variety of cognitive strategies to generate questions, while LLMs struggle to do so. According to the voting results in Figure~\ref{fig3}, the questions generated by our method are the most preferred by users, indicating that our method is closer to human-level question generation compared to LLMs.

\subsection{Ablation Analysis}
\begin{table}[t]%\normalsize
  % \scriptsize
  \centering
   \resizebox{\columnwidth}{!} % 表格长高
  { 
  \renewcommand{\arraystretch}{1.2} % 行间距
  \setlength\tabcolsep{3pt} % 列间距
      \begin{tabular}{cccc}
        \hline
         & $dis(Wiki\_k, q)$ & $dis(Wiki\_k, fq)$ & $dis(q, fq)$ \\
        \hline
        \large w/o re-ranker & \large 32.52 & \large 50.91 & \large 46.86 \\
        \large w/o KGselection & \large 24.11 & \large 49.93 & \large 37.22 \\
        \large w/o llmknowledge & \large 31.85 & \large 61.29 & \large 33.84 \\
        \large Ours & \large 32.85 & \large 55.35 & \large 40.33 \\
        \hline
    \end{tabular}
  }
   \caption{Results of ablation experiments. $Wiki\_k$ denotes the Wiki knowledge output from the \emph{Selection} module; $q$ denotes the input initial question; $fq$ denotes the output follow-up question.}
    \label{table4}
\end{table}

In order to explore the role of each module in generating the final question, we measure the influence of each module by calculating the semantic distance. We use an embedding model to encode the vectors of the Wiki knowledge output by the \emph{Selection} module, the initial question, and the follow-up question, and then compute the pairwise semantic distances. The results are shown in Table~\ref{table4}.

When removing re-ranker to use search engine result directly, both $dis(Wiki_k, q)$ and $dis(Wiki_k, fq)$ decrease compared to the standard framework $(-1.00\%; -8.02\%)$, indicating that inaccurate recognition of the contextual topic indirectly reduces the relevance of the introduced external knowledge to the context. The increase in $dis(q, fq)$ also confirms that the introduction of irrelevant external knowledge causes the follow-up question to rely more on expressions from the original context, thereby reducing its creativity. When we randomly select nodes from the KG, the decreases in $dis(Wiki_k, q)$ and $dis(Wiki_k, fq)$ are even more significant ($-26.60\%; -9.79\%$), demonstrating that selecting node based on both node importance and semantic similarity ensures the relevance of the introduced external knowledge, confirming the necessity of this measure.

When using Wiki knowledge to generate a follow-up question without knowledge fusion operation, we observe a significant increase in $dis(Wiki_k, fq)$ (+10.73\%) while $dis(q, fq)$ decreases sharply (-16.09\%). It suggests that in this case, the generated follow-up question tends to focus more on the Wiki knowledge, which is undesirable. The focus of the question should always revolve around the original context, and the introduced  external knowledge should play an inspiring role rather than becoming the focus of the question. It also shows that allowing LLM to continue writing Wiki knowledge according to context can indeed enhance the contextual relevance of the question while integrating multi-source knowledge.

\begin{table}[t]%\normalsize
  % \scriptsize
  \centering
   \resizebox{\columnwidth}{!} % 表格长高
  { 
  \renewcommand{\arraystretch}{1.2} % 行间距
  \setlength\tabcolsep{2.5pt} % 列间距
      \begin{tabular}{ccccccc}
        \hline
         & $\beta=0$ & $\beta=0.5$ & $\beta=1$ & $\beta=1.5$ & $\beta=2$ & $\beta=\infty$ \\
        \hline
        \large BLEU-1 & \large 11.30 & \large 11.37 & \large \textbf{12.28} & \large 11.46 & \large 11.57 & \large 11.65 \\
        \large BLEU-2 & \large 2.83 & \large 2.97 & \large \textbf{3.26} & \large 3.09 & \large 3.06 & \large 3.24 \\
        \large Perplexity & \large 34.95 & \large 34.64 & \large \textbf{33.74} & \large 35.14 & \large 33.79 & \large 34.17 \\
        \large Topic Consistency & \large 50.19 & \large 50.88 & \large 50.80 & \large \textbf{51.32} & \large 50.36 & \large 50.51 \\
        \hline
    \end{tabular}
  }
   \caption{Compare the generation performance under different weighting parameters $\beta$. We compare the performance by evaluating the BLEU, Perplexity, and Topic Consistency of the final generated question.}
    \label{table5}
\end{table}

\begin{table*}[!t]
  \centering
  \begin{tabular}{|p{7.5cm}|p{8cm}|}
    \hline
    \multicolumn{2}{|p{15.5cm}|}{\textit{\textbf{Initial question}}: Why is the speed of sound constant?  \newline
            \textit{\textbf{Answer}}: The speed of sound is not constant. It depends on the temperature of the medium (and indeed what the medium is made of). It's mathematical formula is square root (specific heat ratio × gas constant × temperature). Loudness is a measure of intensity not speed. Being louder doesn't mean you're heard quicker, it means you're heard more prominently.
            } \\ \hline
    \textit{\textbf{follow-up question}} \newline \textbf{Ours}: Why does temperature affect the speed of sound, is it because of the thermal motion of molecules or something else? For example, what happens to the speed of sound in a medium when the temperature drops from 20°C to below 0°C? \newline
    
    \textbf{BART}: So is the speed of sound constant? \newline
    \textbf{LLaMA3}: What are the factors that affect the speed of sound in a medium? \newline
    \textbf{Qwen2}: How does temperature affect the speed of sound in different mediums? \newline
    \textbf{ChatGLM4}: How does the change in temperature affect the speed of sound in a specific medium? \newline
    \textbf{gpt-3.5-turbo}: How does the temperature of a medium affect the speed of sound? \newline
    & \setlength{\baselineskip}{0.95\baselineskip}\textbf{Related knowledge}: The speed of sound is the distance travelled per unit of time by a sound wave as it propagates through an elastic medium. At 20°C (68°F), the speed of sound in air is about 343 m/s. It depends strongly on temperature as well as the medium through which a sound wave is propagating. At 0°C (32°F), the speed of sound in air is about 331 m/s. Furthermore, the speed of sound is generally faster in solids and liquids than in gases. This is because the particles in solids and liquids are closer together, allowing vibrations to travel more quickly. In addition, the speed of sound also varies depending on the properties of the medium, such as its density and elasticity. As the temperature of a medium increases, the speed of sound generally increases as well. This is due to the fact that the particles in the medium gain more kinetic energy, resulting in faster movement and faster transmission of sound waves. \\ \hline
  \end{tabular}
  \caption{Examples of model-generated follow-up questions. Our method generates questions that incorporate common sense and examples, while baselines cannot.}
  \label{table6}
\end{table*}
\subsection{The effect of $\beta$-value on Selection performance}
We use the BLEU~\cite{ref34}, Perplexity, and Topic Consistency between the Wiki knowledge and the context to evaluate the quality of the Wiki knowledge, as shown in Table ~\ref{table5}. BLEU-1 and BLEU-2 both reach their maximum values when $\beta=1$, then slightly decline as $\beta$ increases. Topic Consistency reaches its maximum at $\beta=1.5$, but higher $\beta$ values lead the model to favor external knowledge based on semantic similarity. Over-reliance on semantics may cause the generated question to deviate from the most relevant core content, thus Topic Consistency decreases slightly at higher $\beta$ values. Perplexity is optimal at $\beta=1$, and either too large or too small a value of $\beta$ results in a loss of semantic fluency of the selected external knowledge relative to the context. In summary, we think that the performance of the \emph{Selection} module is best when $\beta=1$, indicating that node importance and semantic similarity are equally important when selecting external knowledge.

\subsection{Case Study}
In Table~\ref{table6} we present the results of different models for a given context.  We can see that the questions generated by BART essentially rephrase the initial question and do not provide any new information. The questions generated by LLMs are all standard special questions. Although the language style is formal, the questions are rather mechanical and lack the creativity typical of human inquiry. In contrast, while the question generated by our method are consistent with the contents of LLM, they are all aimed at how temperature affects the speed of sound, whereas our question involves factors not mentioned in the context, such as the thermal motion of molecules, and attempts to ask the machine to provide examples for explanation. In the Related knowledge, "the particles in the medium gain more kinetic energy" serves as the knowledge source for "the thermal motion of molecules" in the generated question, and the question also mimics the example in the Related knowledge, asking how the speed of sound changes from 20°C to 0°C. It is clear that the integration of external knowledge helps the model to generate question that is closer to the human level. More examples of the generation of follow-up question can be found in the Appendix.~\ref{appendix2}.

\section{Conclusion}
In this paper, we propose a method to improve follow-up question generation by introducing external knowledge through KG and LLM. Our framework identifies key contextual information, constructs a KG online to acquire background knowledge relevant to the context, and finally integrates multi-source knowledge to generate the follow-up question. Extensive experiments demonstrate that our method outperforms baseline models in both quantitative and qualitative evaluations, generating questions that are richer in information, higher in cognitive complexity, and conducive to moving the conversation to deeper levels.

\section{Limitations}
Although the proposed method achieves remarkable results in experiments, it still has several limitations. First, our framework relies on Wikipedia as an external knowledge source. While Wikipedia contains a vast amount of information, it is not the most accurate source of knowledge in some specific vertical domains. Second, as the KG needs to be constructed in real time, the process is time-consuming, potentially limiting its application in a conversational system. How to balance knowledge accuracy and work efficiency is an important direction for further research in the future.

\section*{Acknowledgements}
This work is partially supported by the project ``Key Laboratory of rich-media Digital Publishing Content Organization and Knowledge Service Open Fund-Research on Knowledge-enhanced Training Techinques of Large Language Model'' No. ZD2024-04/01 and funded by Southeast University-China Mobile Research Institute Joint Innovation Center. We thank the Big Data Computing Center of Southeast University for providing the facility support on the numerical calculations in this paper.

% \section*{Acknowledgments}
% This work is supported by the Natural Science Foundation of China (Grant No. U21A20488). We thank the Big Data Computing Center of Southeast University for providing the facility support on the numerical calculations in this paper.

% This document has been adapted by Emily Allaway from the instructions for earlier ACL and NAACL proceedings, including those for NAACL 2024 by Steven Bethard, Ryan Cotterell and Rui Yan,
% ACL 2019 by Douwe Kiela and Ivan Vuli\'{c},
% NAACL 2019 by Stephanie Lukin and Alla Roskovskaya,
% ACL 2018 by Shay Cohen, Kevin Gimpel, and Wei Lu,
% NAACL 2018 by Margaret Mitchell and Stephanie Lukin,
% Bib\TeX{} suggestions for (NA)ACL 2017/2018 from Jason Eisner,
% ACL 2017 by Dan Gildea and Min-Yen Kan,
% NAACL 2017 by Margaret Mitchell,
% ACL 2012 by Maggie Li and Michael White,
% ACL 2010 by Jing-Shin Chang and Philipp Koehn,
% ACL 2008 by Johanna D. Moore, Simone Teufel, James Allan, and Sadaoki Furui,
% ACL 2005 by Hwee Tou Ng and Kemal Oflazer,
% ACL 2002 by Eugene Charniak and Dekang Lin,
% and earlier ACL and EACL formats written by several people, including
% John Chen, Henry S. Thompson and Donald Walker.
% Additional elements were taken from the formatting instructions of the \emph{International Joint Conference on Artificial Intelligence} and the \emph{Conference on Computer Vision and Pattern Recognition}.

% Bibliography entries for the entire Anthology, followed by custom entries
%\bibliography{anthology,custom}
% Custom bibliography entries only
\bibliography{custom}

\newpage
\appendix

\section{Appendix}
\label{sec:appendix}

\subsection{Questionnaire for Human Evaluation }
\label{appendix1}
\begin{table}[h!]
\centering
\resizebox{\columnwidth}{!} % 表格长高
  { 
  \renewcommand{\arraystretch}{1.2} % 行间距
  \setlength\tabcolsep{2.5pt}
    \begin{tabular}{p{0.5cm}p{8.8cm}}
    \hline
    \multicolumn{2}{c}{\textbf{Questionnaire}} \\ \hline
    \textbf{Q1:} & Whether the question reflect these cognitive approaches: counterfactual, analogical, deductive, and inductive reasoning?   \\ 
     & \(\bigcirc\) Yes \hspace{2em} \(\bigcirc\) No  \hspace{3.9cm} \textbf{\textit{Complexity}} \\ \hline
    \textbf{Q2:} & Whether the question is related to the initial question and answer?  \\ 
     & \(\bigcirc\) Yes \hspace{2em} \(\bigcirc\) No   \hspace{4.2cm}  \textbf{\textit{Relevance}}\\ \hline
    \textbf{Q3:} & Whether the question present new information not mentioned in the context?  \\ 
     & \(\bigcirc\) Yes, there's a lot. \hspace{2em} \(\bigcirc\) No \newline \(\bigcirc\) Yes, but a few.  \hspace{3.4cm}\textbf{\textit{Informativeness}} \\ \hline
    \textbf{Q4:} & Which of these questions do you prefer to ask? \\ 
     & \(\bigcirc\) BART \hspace{1em} \hspace{1em} \(\bigcirc\) LLaMA3 \hspace{1em} \(\bigcirc\) Qwen2 \hspace{1em} \newline \(\bigcirc\) ChatGLM4 \hspace{1em} \(\bigcirc\) gpt-3.5-turbo \hspace{1em} \(\bigcirc\) Ours  \textbf{\textit{Preference}}\\ \hline
    \end{tabular}
    }
    \caption{The questionnaire designed for human evaluation, where $\bigcirc$ indicates a single choice.}
    \label{table2}
\end{table}

\newpage
\subsection{Examples of the Generated Questions}
\label{appendix2}
\begin{table}[H]
  \centering
  \begin{tabular}{|p{7.5cm}|p{8cm}|}
    \hline
    \multicolumn{2}{|p{15.5cm}|}{\textit{\textbf{Initial question}}: Where does ocean trash go?  \newline
            \textit{\textbf{Answer}}: Mostly landfilled.  Some plastics may be sorted out and recycled but this isn't really cost effective and few industries have much use for trash-grade reclaimed ocean plastic.  While landfilling isn't a perfect solution, it is very effective at sacrificing a very tiny area to keep the rest clean.
            } \\ \hline
    \textit{\textbf{follow-up question}} \newline \textbf{Ours}: Why is it not cost-effective to recycle trash-grade ocean plastic, and what alternatives exist? \newline
    
    \textbf{BART}: What about landfilling? \newline
    \textbf{LLaMA3}: What are the main challenges in recycling ocean plastic? \newline
    \textbf{Qwen2}: Why isn't recycling reclaimed ocean plastic cost-effective for most industries? \newline
    \textbf{ChatGLM4}: What are the challenges in recycling ocean plastics, and how can these be addressed? \newline
    \textbf{gpt-3.5-turbo}: What are the economic challenges of recycling ocean plastic? \newline
    & \setlength{\baselineskip}{\baselineskip}\textbf{Related knowledge}: \newline Plastic pollution is the accumulation of plastic objects and particles in the Earth's environment that adversely affects humans, wildlife and their habitats. Plastics that act as pollutants are categorized by size into micro-, meso-, or macro debris. Plastics are inexpensive and durable, making them very adaptable for different uses; as a result, manufacturers choose to use plastic over other materials. However, the chemical structure of most plastics renders them resistant to many natural processes of degradation and decomposition. Consequently, when plastic waste is not properly managed, it often ends up in our oceans. \\ \hline
  \end{tabular}
  \onecolumn \caption{Examples of model-generated follow-up questions.}
  \label{table7}
\end{table}

\begin{table}[H]
  \centering
  \begin{tabular}{|p{7.5cm}|p{8cm}|}
    \hline
    \multicolumn{2}{|p{15.5cm}|}{\textit{\textbf{Initial question}}: Eye colors warmth?  \newline
            \textit{\textbf{Answer}}: Warmth is about color tone.  Warm tones are yellows, reds, oranges. When you have solid brown and add some yellow,  you get a warm brown.  Cool tones are blues and purples, take solid brown and had some blue,  you have cool tone brown eyes.
            } \\ \hline
    \textit{\textbf{follow-up question}} \newline \textbf{Ours}: How does the scattering of light in the iris affect the perception of warmth in eye colors, similar to how light interacts with different surfaces? \newline
    
    \textbf{BART}: Cool tones are blues and purples, take solid brown and had some blue,  you have cool tone brown eyes. \newline
    \textbf{LLaMA3}: What is the effect of adding green to a brown eye color? \newline
    \textbf{Qwen2}: Can you explain how the addition of color tones affects the perception of warmth or coolness in eye colors? \newline
    \textbf{ChatGLM4}: What are some factors that can influence the warmth of eye colors? \newline
    \textbf{gpt-3.5-turbo}: What eye colors are considered warm or cool apart from brown? \newline
    & \setlength{\baselineskip}{\baselineskip}\textbf{Related knowledge}: \newline Eye color is a polygenic phenotypic trait determined by two factors: the pigmentation of the eye's iris and the frequency-dependence of the scattering of light by the turbid medium in the stroma of the iris. The intensity and warmth of eye color can vary depending on the amount of pigmentation in the iris. Warm eye colors, such as shades of brown with yellow undertones, tend to have a rich and vibrant appearance. On the other hand, cool eye colors, like blues and purples, have a more subdued and cooler tone. This variation in eye color is influenced by the interplay of genes and the scattering of light in the iris. \\ \hline
  \end{tabular}
  \caption{Examples of model-generated follow-up questions.}
  \label{table8}
\end{table}

\begin{table}[t]
  \centering
  \begin{tabular}{|p{7.5cm}|p{8cm}|}
    \hline
    \multicolumn{2}{|p{15.5cm}|}{\textit{\textbf{Initial question}}: Can muscle fatigue be bypassed?  \newline
            \textit{\textbf{Answer}}: Having used the fad electro muscle stimulator machines I would say unless the person suffered from paralysis in the limb you were stimulating or was quadriplegic if you were talking full body stimulation, lactic acid would still build up and hurt. Maybe to the point of absolute agony. The only way you can add any extra stamina to muscles with less fatigue is with chemicals, that's why athletes are not allowed to use stimulants like amphetamines because methamphetamine can allow you to fully assert your strengths for longer than it's usually naturally possible but not by a huge amount.  Muscles fibers are essentially chemically induced into an electro magnetic mechanical contraction, which was ordered by a bio-electric nerve signal from the cerebellum for every movement and twitch.
            } \\ \hline
    \textit{\textbf{follow-up question}} \newline \textbf{Ours}: If muscle fatigue is primarily influenced by lactic acid buildup, how do central nervous system factors like neurotransmitter changes also contribute to the sensation of fatigue? \newline
    
    \textbf{BART}: I see, but what if the person was paralyzed? \newline
    \textbf{LLaMA3}: Can bio-electric nerve signals be manipulated to increase muscle strength and endurance? \newline
    \textbf{Qwen2}: What role do chemicals play in reducing muscle fatigue and how are they regulated in sports? \newline
    \textbf{ChatGLM4}: What are the potential risks and side effects of using stimulants, like amphetamines, to enhance athletic performance? \newline
    \textbf{gpt-3.5-turbo}: How do different types of exercise impact muscle fatigue differently? \newline
    & \setlength{\baselineskip}{\baselineskip}\textbf{Related knowledge}: \newline Central nervous system fatigue, or central fatigue, is a form of fatigue that is associated with changes in the synaptic concentration of neurotransmitters within the central nervous system which affects exercise performance and muscle function and cannot be explained by peripheral factors that affect muscle function. In healthy individuals, central fatigue can occur from prolonged exercise and is associated with neurochemical changes in the brain, involving serotonin (5-HT), noradrenaline, and dopamine. These neurotransmitters play a crucial role in regulating mood, motivation, and perception of effort, all of which can influence muscle fatigue. \\ \hline
  \end{tabular}
  \caption{Examples of model-generated follow-up questions.}
  \label{table8}
\end{table}
\end{document}